\documentclass[conference]{IEEEtran}
\IEEEoverridecommandlockouts
\usepackage{cite}
\usepackage{amsmath,amssymb,amsfonts}
\usepackage{algorithmic}
\usepackage{graphicx}
\usepackage{textcomp}
\usepackage{xcolor}
\def\BibTeX{{\rm B\kern-.05em{\sc i\kern-.025em b}\kern-.08em
    T\kern-.1667em\lower.7ex\hbox{E}\kern-.125emX}}

\usepackage[ruled, vlined, linesnumbered]{algorithm2e}
\usepackage{booktabs}
\usepackage{multirow}
\usepackage[caption=false]{subfig}

\def\A{\mbox{\bf{A}}}

\def\HH{\mbox{\bf{H}}}

\def\LL{\mathcal{L}}
\def\M{\mathcal{M}}

\def\SS{\mathcal{S}}

\def\X{\mbox{\bf{X}}}
\def\Y{\mbox{\bf{Y}}}
\def\Z{\mbox{\bf{Z}}}
\def\h{\boldsymbol{h}}
\def\m{\boldsymbol{m}}
\def\w{\boldsymbol{w}}
\def\x{\boldsymbol{x}}
\def\y{\boldsymbol{y}}
\def\z{\boldsymbol{z}}
\def\mmu{\boldsymbol{\mu}}
\def\sig{\boldsymbol{\sigma}}
\def\CE{\text{CE}}
\def\FFN{\text{FFN}}
\def\GCN{\text{GCN}}
\def\KL{\text{KL}}

\def\MSE{\text{MSE}}
\def\Ber{\text{Bernoulli}}
\def\Nor{\text{Normal}}
\def\diag{\text{diag}}
\def\softmax{\text{softmax}}

\begin{document}

\title{Variational Graph Auto-Encoder Based Inductive Learning Method for Semi-Supervised Classification
\thanks{*Corresponding author.}
}
\author{\IEEEauthorblockN{Hanxuan Yang\textsuperscript{1,2}, Zhaoxin Yu\textsuperscript{2,1}, Qingchao Kong\textsuperscript{2,1*}, Wei Liu\textsuperscript{3}, Wenji Mao\textsuperscript{2,1}} \\
\IEEEauthorblockA{\textit{\textsuperscript{1}School of Artificial Intelligence, University of Chinese Academy of Sciences, Beijing, China}\\
\textit{\textsuperscript{2}State Key Laboratory for Multimodal Artificial Intelligence Systems, Institute of Automation,}\\
\textit{Chinese Academy of Sciences, Beijing, China} \\
\textit{\textsuperscript{3}Marketing Service Center of State Grid Zhejiang Electric Power Co. Ltd., Hangzhou, China} \\
\{yanghanxuan2020, yuzhaoxin2024, qingchao.kong, wenji.mao\}@ia.ac.cn, lwei0714@163.com}
}

\maketitle

\begin{abstract}
Graph representation learning is a fundamental research issue in various domains of applications, of which the inductive learning problem is particularly challenging as it requires models to generalize to unseen graph structures during inference. In recent years, graph neural networks (GNNs) have emerged as powerful graph models for inductive learning tasks such as node classification, whereas they typically heavily rely on the annotated nodes under a fully supervised training setting. Compared with the GNN-based methods, variational graph auto-encoders (VGAEs) are known to be more generalizable to capture the internal structural information of graphs independent of node labels and have achieved prominent performance on multiple unsupervised learning tasks. However, so far there is still a lack of work focusing on leveraging the VGAE framework for inductive learning, due to the difficulties in training the model in a supervised manner and avoiding over-fitting the proximity information of graphs. To solve these problems and improve the model performance of VGAEs for inductive graph representation learning, in this work, we propose the Self-Label Augmented VGAE model. To leverage the label information for training, our model takes node labels as one-hot encoded inputs and then performs label reconstruction in model training. To overcome the scarcity problem of node labels for semi-supervised settings, we further propose the Self-Label Augmentation Method (SLAM), which uses pseudo labels generated by our model with a node-wise masking approach to enhance the label information. Experiments on benchmark inductive learning graph datasets verify that our proposed model archives promising results on node classification with particular superiority under semi-supervised learning settings.
\end{abstract}


\section{Introduction}

Graph representation learning aims to learn low-dimensional embeddings of labeled graph nodes and has become a critical problem with plenty of applications in real-world scenarios, represented by the node classification task. Learning graph representations requires a model to leverage both the general structural information of the whole graph and the specific features and label information of each node. Typically, the graph representation learning problem can be divided into transductive and inductive learning. Compared to the standard transductive learning setting where all nodes are visible during both the training and testing processes, the \textit{inductive learning} problem assumes the testing nodes (and their attribute features and related edges) to be unseen during training and thus is more challenging for graph models to generalize to unknown graph structures \cite{hamilton2017inductive}. Classifying unseen nodes under the inductive learning setting is very prevalent and important in many real-world graph structures, such as the dynamic evolving networks \cite{leskovec2005graphs, paranjape2017motifs, weber2019anti} and cross-graph networks \cite{subramanian2005gene, rozemberczki2019gemsec}.


With the development of deep learning, graph neural networks (GNNs) have emerged as powerful graph representation learning methods \cite{kipf2017semi, hamilton2017inductive, velivckovic2018graph, xu2020inductive, zeng2020graphsaint, ciano2022inductive, huang2022graph, anghinoni2023transgnn, cavallo2023gcnh}. However, existing GNN-based methods heavily rely on plenty of annotated data for training and only consider the fully supervised inductive learning setting with all visible nodes labeled. This can severely constrain the practical applications of these methods, since the label information of many nodes can be unavailable due to the data incompleteness or expensive annotating cost in real-world scenarios, i.e., the semi-supervised inductive learning setting.


Compared with the GNN-based methods, the variational graph auto-encoder (VGAE)  \cite{kipf2016variational} based generative graph models are known to be more generalizable for capturing the underlying proximity information and have shown promising performance on multiple unsupervised graph learning tasks \cite{grover2019graphite, mehta2019stochastic, sarkar2020graph, tan2022mgae, hou2022graphmae, guo2022graph, fan2022heterogeneous, li2023maskgae}. These methods typically benefit from the good generalizability of variational auto-encoders (VAEs) \cite{kingma2013auto} by adding the Kullback-Leibler (KL) divergence as a data-agnostic regularization term to the loss function \cite{takahashi2019variational}, and thus can effectively alleviate the overreliance of GNNs on data annotations. However, currently there is still a lack of research attempting to leverage VGAEs for inductive graph representation learning. One of the primary challenges is how to leverage the label information of graph nodes under the unsupervised training paradigm of VGAEs, especially for scarce annotation scenarios. Existing VGAE-based methods typically first learn node embeddings in an unsupervised learning manner and then use node labels to train an additional classifier on top of the learned embeddings \cite{tan2022mgae, hou2022graphmae, li2023maskgae}, which may significantly increase the training burden. In addition, as revealed by \cite{velivckovic2018deep, hassani2020contrastive}, VGAEs tend to over-fit the proximity information of graph structures, which can also hurt the model performance for node classification.

To improve the performance of VGAEs for semi-supervised inductive graph representation learning, we propose the Self-Label Augmented Variational Graph Auto-Encoder (SLA-VGAE) model. Our model consists of a graph convolutional network (GCN) \cite{kipf2017semi} encoder to perform neighbor aggregation and a novel label reconstruction decoder for model training. To better leverage the label information within the VGAE framework, we encode the node labels as one-hot features and then employ the decoder to reconstruct the labels instead of the adjacency matrix. In addition, to deal with the scarcity problem of node labels under the semi-supervised learning setting, we propose a Self-Label Augmentation Method (SLAM) to generate pseudo node labels with our model using a node-wise masking approach, which can also enhance the model generalizability for inferring the representations of unseen nodes. We conduct extensive experiments on the inductive learning graph datasets of node classification. The results verify that our proposed model can significantly improve the performance of VGAEs for semi-supervised graph learning and achieve superior or comparable results to the state-of-the-art methods.

The main contributions of our work are as follows:
\begin{itemize}
        \item We develop a VGAE-based inductive learning method for semi-supervised node classification with a novel label reconstruction decoder to reconstruct node labels instead of adjacency matrices for training.
	\item To address the scarcity problems of node labels and boost the model generalizability for inductive learning, we propose a Self-Label Augmentation Method (SLAM) to generate pseudo labels using a node-wise masking approach.
	\item Experimental results on inductive learning graph datasets verify that our model achieves promising performance for node classification with particular superiority under semi-supervised settings.
\end{itemize}

\section{Related Work}

In this section, we briefly review the representative work related to inductive graph representation learning and the VGAE-based graph models.

\subsection{Inductive Graph Representation Learning}
Inductive graph representation learning aims to learn low-dimensional node embeddings based on the graph topology and label information, where the nodes for inference are unseen during the training process. Representative graph models for inductive learning are typically based on the GNN framework. These methods learn node representations by repeatedly performing neighbor aggregation based on graph topology, and can be applied to model variable graph structures for inductive learning \cite{kipf2017semi, hamilton2017inductive, velivckovic2018graph, xu2020inductive, zeng2020graphsaint, ciano2022inductive, huang2022graph, anghinoni2023transgnn, cavallo2023gcnh}. To further improve the model performance, some recent work proposes the label propagation method to combine node labels with attribute features as model input \cite{zhang2022graph}, and enhance the label information with pseudo node labels generated by a pre-trained teacher model \cite{sun2021scalable, zhang2021scr}. These methods have achieved prominent performance for semi-supervised node classification under transductive learning settings, but the more challenging inductive learning problem with scarce label information remains to be investigated.

\subsection{Variational Graph Auto-Encoders}
The VGAE-based methods are probabilistic models for graph representation learning. These methods generate latent variables as node embeddings and perform graph reconstruction for model training. Taking the KL divergence as a regularization term, the VGAE-based methods benefit from good generalizability and have achieved promising results on unsupervised learning tasks such as link prediction and community detection \cite{kipf2016variational, grover2019graphite, mehta2019stochastic, sarkar2020graph, guo2022graph, fan2022heterogeneous}. Nevertheless, the VGAE-based methods typically show poor performance on supervised or semi-supervised learning tasks such as node classification, as they tend to over-fit the internal graph proximity and cannot fully leverage the external label information of graph datasets. Recently, some work \cite{tan2022mgae, hou2022graphmae, li2023maskgae} attempts to improve the learning power of VGAEs beyond link prediction using masking approaches. For example, GraphMAE \cite{hou2022graphmae} randomly masks the attribute features of some nodes and then reconstructs the node features. MaskGAE \cite{li2023maskgae} masks some paths or edges of a graph and reconstructs the adjacency matrix as well as node degrees for training. However, none of these mask approaches can adapt to the variable graph structures for inductive learning, where some nodes are completely unseen (including the node attribute features and proximity structures). Moreover, most of the existing VGAE-based methods employ a non-end-to-end training manner for supervised learning tasks and must train an additional classifier for classification, which can also impact the model performance on these tasks.

To this end, we develop the VGAE framework for semi-supervised graph representation learning by reconstructing the node labels, instead of the adjacency matrix. In addition, we also leverage a node-wise masking approach to generate pseudo node labels with some nodes randomly masked, so as to adapt to the scarcity of ground-truth node labels and further improve the model generalizability for inductive learning.

\begin{figure*}[t]
	\begin{center}
		\includegraphics[width=\textwidth]{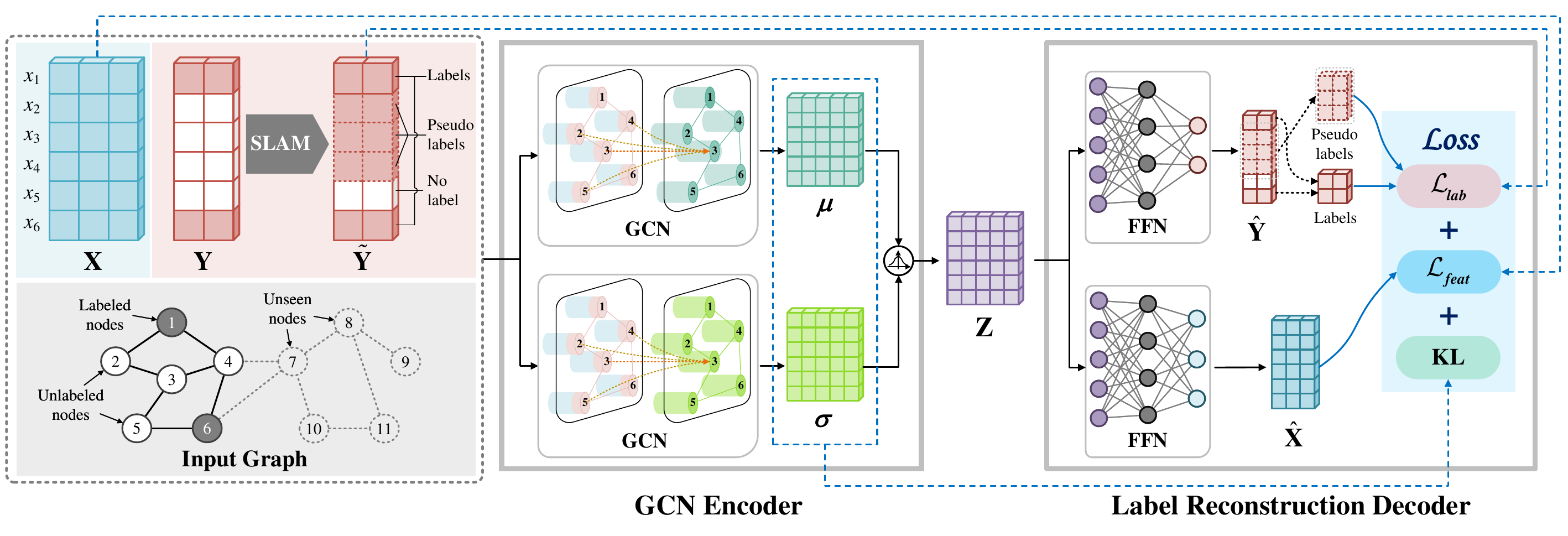}\par
		\caption{The sketch of our proposed SLA-VGAE model. During training, the nodes for testing and validation are unseen in the input graph. The true node labels are augmented via SLAM (after the warm-up stage) and then combined with node features as input of the GCN encoder to generate node representations. The decoder reconstructs the augmented node labels and features and calculates the loss function for model training (blue dashed arrows).}
		\label{fig:model}
	\end{center}
\end{figure*}

\section{Method}

We propose the Self-Label Augmented Variational Graph Auto-Encoder (SLA-VGAE) for semi-supervised graph representation learning. Our model consists of an encoder that employs GCN layers to learn node embeddings, and a decoder that reconstructs node labels as well as attribute features for model training. The overall framework of our model is presented in Fig.~\ref{fig:model}.

\subsection{GCN Encoder}

The encoder employs GCN layers to perform neighbor aggregation to learn node embeddings. Given an adjacency matrix of a graph with $n$ nodes $\A\in\{0,1\}^{n\times n}$, the GCN embeddings $\HH^{(l)}=(\h_1^{(l)},\dots,\h_n^{(l)})'$ of the $l$-th layer, $l=1,\dots,L$, are obtained as
\begin{align}\label{gcn}
	\HH^{(l)}=\GCN^{(l)}(\A, \HH^{(l-1)}).
\end{align}
The initial input $\HH^{(0)}$ is defined as a combination of the node attribute features (if available) $\X=(\x_1,\dots,\x_n)'$ and one-hot encoded node labels $\Y=(\y_1,\dots,\y_n)'$ (the unlabeled and testing nodes are encoded as zero vectors). In addition, to enhance the label information for the unlabeled data, we propose a label augmentation method using a node-wise masking approach, which we shall elaborate in Section~\ref{sec:slam}. Therefore, the input features can be formulated as
\begin{align}
	\HH^{(0)}=[\X\vert\tilde{\Y}],
\end{align}
where $[\cdot\vert\cdot]$ indicates the concatenation operation and $\tilde{\Y}$ are the augmented node labels.

The GCN embeddings are then leveraged as variational parameters to generate Normal latent variables as node representations via Monte Carlo (MC) sampling. Formally, the node representations $\Z=(\z_1,\dots,\z_n)'$ is generated as, for $i=1,\dots,n$,
\begin{align}
	\z_i\sim\Nor(\mmu_i,\diag(\sig_i^2)),
\end{align}
where the mean $\mmu_i$ and standard deviation $\sig_i$ parameters are obtained from the GCN output layer ($l=L$). Following the vanilla VGAEs, the reparameterization trick is adopted for gradient optimization \cite{kingma2013auto}.

\begin{algorithm}[tb]
	\caption{Training SLA-VGAE}\label{train}
	\KwIn{Graph adjacency matrix $\A$; node features $\X$; node labels $\Y$}
	\KwOut{Predicted node labels $\hat{\Y}$}
	\LinesNumbered
	\textbf{Initialize} weight parameters $\w$ of model $\M$;\\
	\For{$t=1,\dots,T$}{
		\If{$t<=t_{warm-up}$}{
			$\tilde{\Y}=\Y$;\\
		}
		\Else{
			$\tilde{\Y}=\textsc{SLAM}(\A,\X,\Y)$;\\
		}
		$\HH^{(0)}=[\X\vert\tilde{\Y}]$;\\
		\For{$l=1,\dots,L-1$}{
			$\HH^{(l)}=\textsc{GCN}^{(l)}(\A,\HH^{(l-1)})$;\\
		}
		$\mmu=\textsc{GCN}_{\mu}^{(L)}(\HH^{(L-1)})$;\\
		$\sig=\textsc{GCN}_{\sigma}^{(L)}(\HH^{(L-1)})$;\\
		$\Z=\textsc{NormalSampling}(\mmu,\sig)$;\\
		$\hat{\Y}=\textsc{Softmax}(\textsc{FFN}_y(\Z))$;\\
		$\hat{\X}=\textsc{FFN}_x(\Z)$;\\
		$\LL_{lab}=\textsc{CE}(\tilde{\Y},\hat{\Y})$;\\
		$\LL_{feat}=\textsc{MSE}(\X,\hat{\X})$;\\
		$\LL=\LL_{lab}+\lambda_{feat}\LL_{feat}+\KL[q(\Z)\vert p(\Z)]$;\\
		$\w\gets\w-\eta\nabla_{\omega}\mathcal{L}$;\\
	}
\end{algorithm}

\subsection{Lable Reconstruction Decoder}
To leverage the label information for supervised model training, we propose the label reconstruction decoder that reconstructs node labels as well as attribute features using feedforward networks (FFNs), i.e.,
\begin{align}
	\hat{\Y}&=\softmax(\FFN_y(\Z)),\\
	\hat{\X}&=\FFN_x(\Z).
\end{align}

The loss function of our model is defined as a combination of the reconstruction loss and the KL divergence between the variational posterior and prior distributions of node representations. Specifically, the reconstruction loss contains reconstructing the node labels and features, i.e.,
\begin{align}\label{loss}
	\LL_{lab}&=\CE(\tilde{\Y},\hat{\Y}),\\
	\LL_{feat}&=\MSE(\X,\hat{\X}),
\end{align}
where $\CE$ and $\MSE$ indicate the cross entropy and mean square error, respectively. Note that we only calculate the label reconstruction loss $\LL_{lab}$ of the labeled nodes for gradient optimization, and the unlabeled nodes (including those in the testing and validation sets) are excluded for calculating $\LL_{lab}$. Finally, the full loss function is given as
\begin{align}
	\LL=\LL_{lab}+\lambda_{feat}\LL_{feat}+\KL[q(\Z)\vert p(\Z)],
\end{align}
where $\lambda_{feat}$ is a tuning hyperparameter, $q(\cdot)$ and $p(\cdot)$ denote the variational posterior and the standard Normal prior of node representations, respectively. The pseudo code for training our model is given in Algorithm~\ref{train}.

\begin{figure*}[t]
	\begin{center}
		\includegraphics[width=\textwidth]{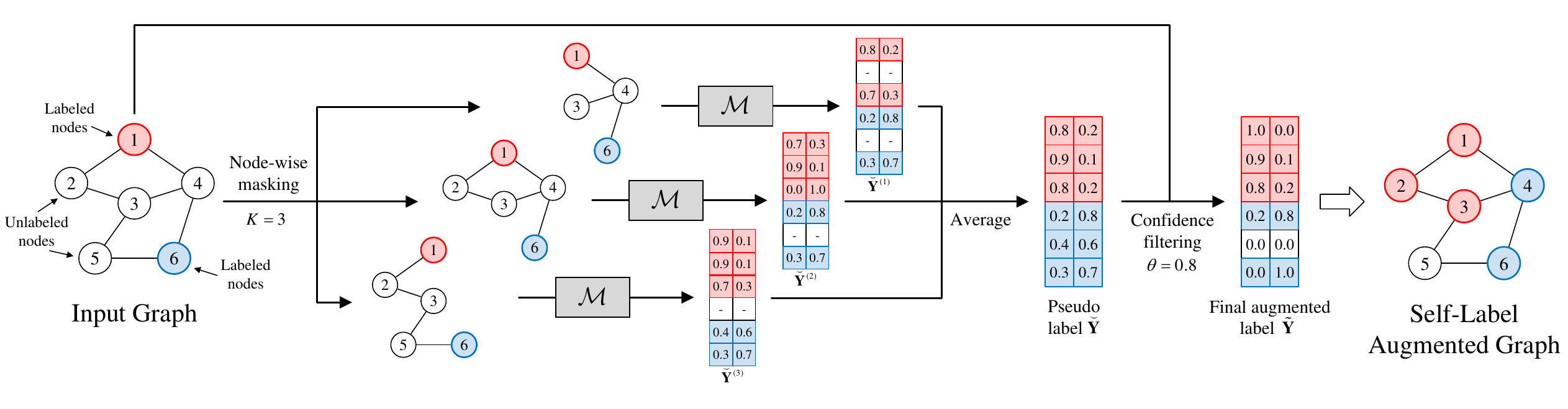}\par
		\caption{An illustration of the proposed SLAM for label augmentation. The input graph is randomly masked with some nodes and fed into the model $\M$ obtained from the last iteration of training to generate labels of the unmasked nodes. The final augmented labels $\tilde{\Y}$ are computed by averaging over all generated labels and then filtering the low-confident samples, where the ground-truth labels of the labeled nodes are retained as well.}
		\label{fig:slam}
	\end{center}
\end{figure*}

\subsection{Self-Label Augmentation Method}\label{sec:slam}

Our proposed model leverages node labels as input features to perform label reconstruction. However, in practice, many nodes are unlabeled, making the label features very sparse and insufficient for model training. To deal with this issue, we propose a Self-Label Augmentation Method (SLAM) to enhance the label information by generating pseudo labels using the model itself, which are then leveraged as augmented labels at the next iteration after confidence filtering. The procedure of SLAM is illustrated in Fig.~\ref{fig:slam}

The complete training process is divided into two stages. During the first stage, referred to as the warm-up stage, we only use the true labels to train the model. Then, after several iterations, the training process enters the second stage, when we add the pseudo labels generated by the model $\M$ obtained from the last training iteration with all weight parameters frozen. Specifically, we generate node labels $\breve{\Y}^{(k)}$ using the model for $K$ times, $k=1,\dots,K$, and the pseudo labels are obtained as the averages of all generated labels, i.e.,
\begin{align}
 	\breve{\Y}=\frac{1}{K}\sum_{k=1}^{K}\breve{\Y}^{(k)}.
\end{align}

In addition, to ensure high confidence for the generated labels, we set a threshold $\theta$ to filter out the low-confident pseudo labels. Thus, the final augmented node labels $\tilde{\Y}=(\tilde{\y}_1,\dots,\tilde{\y}_n)'$ for model training is formed as, for $i=1,\dots,n$,
\begin{align}
	\tilde{\y}_i=\left\{
	\begin{array}{ll}
		\y_i,&i\in\SS_{tr},\\
		\breve{\y}_i,&i\in\SS\backslash\SS_{tr}\text{ and }\breve{\y}_i>\theta,\\
		\bf{0},&\text{otherwise}.
	\end{array}
	\right.
\end{align}
where $\SS$ and $\SS_{tr}$ denote the full set and training set of the nodes, respectively.

To further improve the model generalizability for inductive learning, where some nodes are unseen in the training graph, we propose a node-wise masking approach to generate the pseudo labels by randomly masking some nodes each time. The node masks $\m^{(k)}=(m_1^{(k)},\dots,m_n^{(k)})'$ are generated via Bernoulli sampling, i.e., for $i=1,\dots,n$,
\begin{align}
    m_i^{(k)}\sim\Ber(p),
\end{align}
where $p$ is the probability for each node to be unmasked. With the node-wise masking approach, our model is facilitated to adapt to variable graph structures during the training process, and thus can be more generalizable for learning representations of graphs with some nodes invisible. The pseudo code of SLAM is provided in Algorithm~\ref{slam}.

\begin{algorithm}[tb]
	\caption{Self-Label Augmentation Method}\label{slam}
	\KwIn{Graph adjacency matrix $\A$; node features $\X$; node labels $\Y$}
	\KwOut{Augmented node labels $\tilde{\Y}$}
	\LinesNumbered 
	\For{$k=1,\dots,K$}{
		$\m^{(k)}=\textsc{BernoulliSampling}(p)$;\\
		$\A_{mask}^{(k)}=\textsc{Mask}(\A,\m^{(k)})$;\\
		$\hat{\Y}^{(k)}=\M(\A_{mask}^{(k)},\X,\Y)$;\\
	}
	$\hat{\Y}=\frac{1}{K}\sum_{k=1}^{K}\hat{\Y}^{(k)}$;\\
	$\tilde{\Y}=\textsc{ConfidenceFiltering}(\hat{\Y})$;\\
\end{algorithm}


\begin{table*}[tb]
	\centering
	\caption{Experimental results of node classification on Flickr with different labeling rates. The best results are in bold and the second-best ones are underlined.}\label{results_flickr}
	\resizebox{\textwidth}{!}{
		\begin{tabular}{cccccccc}
			\toprule
			&&\multicolumn{2}{c}{\textbf{1\%}}&\multicolumn{2}{c}{\textbf{10\%}}&\multicolumn{2}{c}{\textbf{100\%}}\\[1pt]
                &&Accuracy&MCC&Accuracy&MCC&Accuracy&MCC\\
			\midrule
			\multirow{6}{*}{GNN-Based}&GCN&\underline{0.428 $\pm$ 0.003}&\underline{0.168 $\pm$ 0.005}&0.473 $\pm$ 0.002&0.184 $\pm$ 0.004&0.491 $\pm$ 0.001&0.216 $\pm$ 0.001\\[1pt]
			&GraphSAGE&0.343 $\pm$ 0.005&0.139 $\pm$ 0.009&0.411 $\pm$ 0.004&\underline{0.189 $\pm$ 0.004}&0.499 $\pm$ 0.001&0.236 $\pm$ 0.002\\[1pt]
			&GraphSAINT&0.414 $\pm$ 0.011&0.139 $\pm$ 0.007&\underline{0.476 $\pm$ 0.013}&0.156 $\pm$ 0.017&0.504 $\pm$ 0.004&0.245 $\pm$ 0.006\\[1pt]
			&GNN-INCM&0.386 $\pm$ 0.013&0.124 $\pm$ 0.016&0.437 $\pm$ 0.005&0.148 $\pm$ 0.008&0.493 $\pm$ 0.005&0.248 $\pm$ 0.005\\[1pt]
			&GAMLP&0.325 $\pm$ 0.025&0.108 $\pm$ 0.011&0.410 $\pm$ 0.014&0.176 $\pm$ 0.002&\underline{0.505 $\pm$ 0.005}&\underline{0.257 $\pm$ 0.021}\\[1pt]
			&TransGNN&0.382 $\pm$ 0.020&0.117 $\pm$ 0.015&0.425 $\pm$ 0.018&0.157 $\pm$ 0.014&0.488 $\pm$ 0.016&0.241 $\pm$ 0.014\\
			\midrule
			\multirow{3}{*}{VGAE-Based}&VGAE&0.396 $\pm$ 0.006&0.040 $\pm$ 0.013&0.425 $\pm$ 0.008&0.101 $\pm$ 0.007&0.494 $\pm$ 0.005&0.220 $\pm$ 0.003\\[1pt]
			&GraphMAE&0.425 $\pm$ 0.006&0.042 $\pm$ 0.016&0.431 $\pm$ 0.006&0.107 $\pm$ 0.016&0.441 $\pm$ 0.001&0.118 $\pm$ 0.019\\[1pt]
			&MaskGAE&0.427 $\pm$ 0.003&0.052 $\pm$ 0.011&0.455 $\pm$ 0.012&0.142 $\pm$ 0.005&0.496 $\pm$ 0.004&0.224 $\pm$ 0.005\\
			\midrule
			\textbf{Ours}&\textbf{SLA-VGAE}&\textbf{0.475 $\pm$ 0.006}&\textbf{0.195 $\pm$ 0.007}&\textbf{0.488 $\pm$ 0.005}&\textbf{0.208 $\pm$ 0.010}&\textbf{0.507 $\pm$ 0.005}&\textbf{0.259 $\pm$ 0.015}\\
			\bottomrule
		\end{tabular}
	}
\end{table*}


\section{Experiments}

To evaluate the performance of our proposed SLA-VGAE for supervised and semi-supervised graph representation learning, we conduct a series of node classification experiments on benchmark inductive learning graph datasets.

\subsection{Datasets}

We consider two inductive learning social networks, i.e., Flickr \cite{zeng2020graphsaint} and Reddit \cite{hamilton2017inductive}. Flickr is a collection of 800 ego-graphs containing 89,250 images uploaded to a social website as nodes and the common metadata such as locations and tags shared by two images as edges. The images are divided into 7 categories based on their tags. The node features are obtained using bag-of-word embeddings of the image descriptions. These ego-graphs are randomly selected as 50\% for training, 25\% for testing
and 25\% for validation.

Reddit is a dynamic evolving network collected from a news comment website in September 2014, where the nodes represent posts and two nodes are connected if they are commented by the same user. The node features are word vectors of the post titles and comments, and the labels are the post communities. Posts in the first 20 days of the month are used as the training set and others in the last 10 days are randomly selected as 70\% for testing and 30\% for validation.


\subsection{Baselines}
We compare our model with the state-of-the-art methods for graph representation learning, including six GNN-based methods and three VGAE-based methods. GCN \cite{kipf2017semi} first proposes a GNN framework for semi-supervised graph learning by performing neighbor aggregation based on the graph Laplacian. GraphSAGE \cite{hamilton2017inductive} first focuses on the inductive learning problem on graphs and proposes a neighbor sampling method to aggregate neighbor information based on graph topology. GraphSAINT \cite{zeng2020graphsaint} further introduces an efficient graph sampling method for inductive learning by sampling subgraphs instead of nodes or edges. GNN-INCM \cite{huang2022graph} employs embedding clustering and graph reconstruction to deal with the imbalance problem of node classes. GAMLP \cite{zhang2022graph} leverages the label propagation method to improve model performance and is currently among the methods with the best accuracy results for node classification. TransGNN \cite{anghinoni2023transgnn} develops a message-passing technique to perform transductive learning for semi-supervised node classification.

The VGAE-based comparative methods include the vanilla VGAE \cite{kipf2016variational}, which generates latent variables from Normal distributions as node embeddings and reconstructs the adjacency matrix for model training. GraphMAE \cite{hou2022graphmae} randomly masks node features and then reconstructs the input features. MaskGAE \cite{li2023maskgae} randomly masks some edges in a graph to mitigate over-fitting the proximity information. Note that the three VGAE-based methods cannot employ the label information in an end-to-end training manner. Following the standard settings \cite{hou2022graphmae, li2023maskgae}, we fit logistic regression models for node classification on top of the embeddings learned by these methods.

\begin{table*}[tb]
	\centering
	\caption{Experimental results of node classification on Reddit with different labeling rates. The best results are in bold and the second-best ones are underlined.}\label{results_reddit}
	\resizebox{\textwidth}{!}{
		\begin{tabular}{cccccccc}
			\toprule
			&&\multicolumn{2}{c}{\textbf{1\%}}&\multicolumn{2}{c}{\textbf{10\%}}&\multicolumn{2}{c}{\textbf{100\%}}\\[1pt]
                &&Accuracy&MCC&Accuracy&MCC&Accuracy&MCC\\
			\midrule
			\multirow{6}{*}{GNN-Based}&GCN&\underline{0.921 $\pm$ 0.000}&\underline{0.916 $\pm$ 0.001}&0.940 $\pm$ 0.000&0.937 $\pm$ 0.001&0.947 $\pm$ 0.002&0.944 $\pm$ 0.000\\[1pt]
			&GraphSAGE&0.889 $\pm$ 0.001&0.883 $\pm$ 0.001&0.939 $\pm$ 0.000&0.936 $\pm$ 0.002&0.935 $\pm$ 0.004&0.950 $\pm$ 0.004\\[1pt]
			&GraphSAINT&0.660 $\pm$ 0.006&0.644 $\pm$ 0.004&0.916 $\pm$ 0.007&0.911 $\pm$ 0.007&\underline{0.961 $\pm$ 0.003}&\underline{0.958 $\pm$ 0.004}\\[1pt]
			&GNN-INCM&0.862 $\pm$ 0.005&0.854 $\pm$ 0.007&0.931 $\pm$ 0.004&0.935 $\pm$ 0.015&0.942 $\pm$ 0.003&0.949 $\pm$ 0.005\\[1pt]
			&GAMLP&0.846 $\pm$ 0.014&0.839 $\pm$ 0.009&\underline{0.943 $\pm$ 0.009}&\underline{0.937 $\pm$ 0.000}&\textbf{0.967 $\pm$ 0.000}&\textbf{0.965 $\pm$ 0.002}\\[1pt]
			&TransGNN&0.852 $\pm$ 0.014&0.847 $\pm$ 0.012&0.926 $\pm$ 0.019&0.928 $\pm$ 0.011&0.946 $\pm$ 0.010&0.947 $\pm$ 0.006\\
			\midrule
			\multirow{3}{*}{VGAE-Based}&VGAE&0.642 $\pm$ 0.023&0.629 $\pm$ 0.018&0.730 $\pm$ 0.012&0.715 $\pm$ 0.008&0.928 $\pm$ 0.006&0.921 $\pm$ 0.005\\[1pt]
			&GraphMAE&0.918 $\pm$ 0.001&0.914 $\pm$ 0.003&0.932 $\pm$ 0.004&0.934 $\pm$ 0.003&0.955 $\pm$ 0.003&0.952 $\pm$ 0.005\\[1pt]
			&MaskGAE&0.881 $\pm$ 0.004&0.875 $\pm$ 0.003&0.935 $\pm$ 0.000&0.932 $\pm$ 0.000&0.948 $\pm$ 0.002&0.945 $\pm$ 0.002\\
			\midrule
			\textbf{Ours}&\textbf{SLA-VGAE}&\textbf{0.938 $\pm$ 0.001}&\textbf{0.936 $\pm$ 0.001}&\textbf{0.948 $\pm$ 0.000}&\textbf{0.945 $\pm$ 0.000}&0.955 $\pm$ 0.001&0.953 $\pm$ 0.001\\
			\bottomrule
		\end{tabular}
	}
\end{table*}

\subsection{Implementation Details}
To evaluate the performance of our model for inductive graph representation learning under both supervised and semi-supervised settings, we consider three different labeling rates of the training sets. Specifically, for each dataset, we randomly keep 1\%, 10\% and 100\% nodes of the training set with labels, respectively, and the labels of all other nodes are masked during the training process. Following the standard settings of inductive learning, all models are trained on a subgraph of each dataset where the nodes (and their related edges) of the testing and validation sets are unseen, and then tested on the full graph with all nodes visible.

For the hyperparameter settings of our proposed SLA-VGAE, we use two GCN layers with 512 hidden channels each for the encoder, and three fully connected layers with 512 channels each for the decoder. The tuning hyperparameter of feature reconstruction is set as $\lambda_{feat}=0.1$. During training, we first run 1 epoch as the warm-up stage and then leverage the proposed SLAM for label augmentation, of which the hyperparameters are set as the generation times $K\in\{1,2\}$, the unmasking probability $p=0.7$, and the confidential threshold $\theta=0.9$. The learning rate $\eta$ is fixed in $\{0.001,0.005\}$. The comparative methods are implemented with the same number of layers and hidden channels as that of our SLA-VGAE for the encoder, and other hyperparameters are set as default in their released source code. All models are trained for less than 500 epochs with an early-stopping strategy.

\subsection{Result Analysis}
We select the most common classification accuracy and the Matthews correlation coefficient (MCC) \cite{gorodkin2004comparing} as metrics, of which the latter is widely used for evaluating classification on imbalanced data \cite{huang2022graph, anghinoni2023transgnn}. The experimental results on the two datasets are presented in Table~\ref{results_flickr} and \ref{results_reddit}, respectively, where all results are reported based on the means and standard deviations of 5 independent implementations with different random seeds. The experimental results verify that our proposed SLA-VGAE shows significantly superior performance over all comparative methods under the semi-supervised settings, and at least comparable performance under the fully supervised setting. Specifically, as the labeling rate decreases, the comparative methods present significant declines in model performance. For example, the classification accuracy of GAMLP drops about 35.6\% and 12.5\% on Flickr and Reddit, respectively, whereas that of our SLA-VGAE only drops 6.3\% and 1.8\% on the two datasets. It seems that the simple GCN performs more robustly under the weakly supervised settings, since the other more complex comparative methods have more parameters and require more labeled data for training. In contrast, our SLA-VGAE can effectively alleviate the label scarcity problem for parameter optimization via the proposed label augmentation method.

\begin{figure}[tb]
	\centering
        \hspace{-5mm}
	\subfloat[Flickr]{
		\includegraphics[width=0.49\columnwidth]{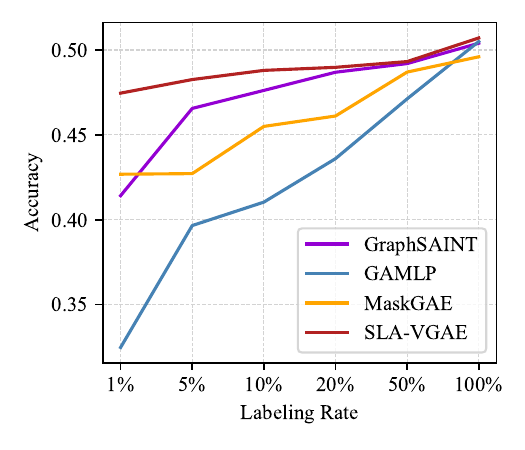}
	}\hspace{-5mm}
	\subfloat[Reddit]{
		\includegraphics[width=0.49\columnwidth]{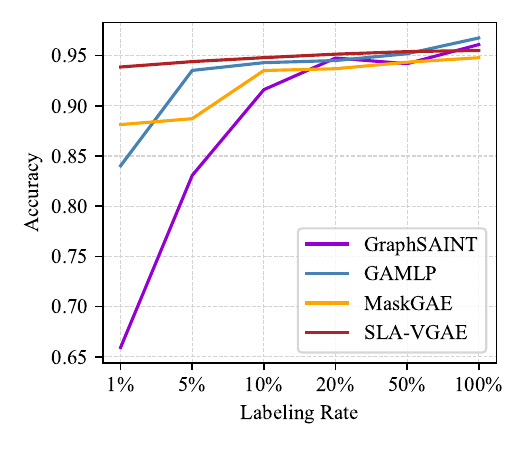}
	}
	\centering
	\caption{Experimental results of node classification accuracy on the inductive learning datasets with different labeling rates.\label{labeling_rate}}
\end{figure}

In addition, we also compare our proposed SLA-VGAE with the three most powerful comparative methods for node classification, i.e., GraphSAINT, GAMLP and MaskGAE, on the two datasets with more scales of labeling rates, as presented in Fig.~\ref{labeling_rate}. The results intuitively demonstrate that the performance of our model is much more robust than the comparative methods under weakly supervised settings with scarce labels, and the superiority of our model consistently grows larger as the labeling rate decreases.

\begin{table}[tb]
	\centering
	\caption{Ablation study results of node classification accuracy on Flickr with different labeling rates. The best results are in bold.}\label{ablation_flickr}
	\resizebox{\columnwidth}{!}{
		\begin{tabular}{lccc}
			\toprule
                &1\%&10\%&100\%\\
			\midrule
			\textbf{SLA-VGAE}&\textbf{0.475 $\pm$ 0.006}&\textbf{0.488 $\pm$ 0.005}&\textbf{0.507 $\pm$ 0.005}\\[1pt]
                \quad w/o feature&0.472 $\pm$ 0.002&0.480 $\pm$ 0.004&0.502 $\pm$ 0.004\\[1pt]
                \quad w/o mask&0.467 $\pm$ 0.010&0.483 $\pm$ 0.005&0.501 $\pm$ 0.000\\[1pt]
                \quad w/o pseudo&0.463 $\pm$ 0.000&0.472 $\pm$ 0.002&0.490 $\pm$ 0.007\\[1pt]
                \quad w/o label&0.454 $\pm$ 0.004&0.471 $\pm$ 0.000&0.488 $\pm$ 0.002\\
			\bottomrule
		\end{tabular}
	}
\end{table}

\begin{table}[tb]
	\centering
	\caption{Ablation study results of node classification accuracy on Reddit with different labeling rates. The best results are in bold.}\label{ablation_reddit}
	\resizebox{\columnwidth}{!}{
		\begin{tabular}{lccc}
			\toprule
                &1\%&10\%&100\%\\
			\midrule
			\textbf{SLA-VGAE}&\textbf{0.938 $\pm$ 0.006}&\textbf{0.948 $\pm$ 0.000}&\textbf{0.955 $\pm$ 0.001}\\[1pt]
                \quad w/o feature&0.933 $\pm$ 0.001&0.936 $\pm$ 0.001&0.951 $\pm$ 0.000\\[1pt]
                \quad w/o mask&0.936 $\pm$ 0.001&0.941 $\pm$ 0.001&0.952 $\pm$ 0.003\\[1pt]
                \quad w/o pseudo&0.922 $\pm$ 0.002&0.935 $\pm$ 0.005&0.951 $\pm$ 0.003\\[1pt]
                \quad w/o label&0.921 $\pm$ 0.001&0.935 $\pm$ 0.001&0.950 $\pm$ 0.001\\
			\bottomrule
		\end{tabular}
	}
\end{table}

\begin{figure*}[htbp]
	\centering
        \hspace{-5mm}
	\subfloat[]{
		\includegraphics[width=0.3\textwidth]{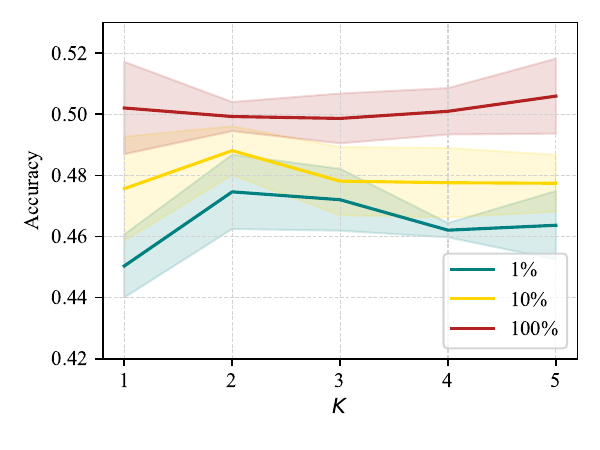}
	}\hspace{-5mm}
	\subfloat[]{
		\includegraphics[width=0.3\textwidth]{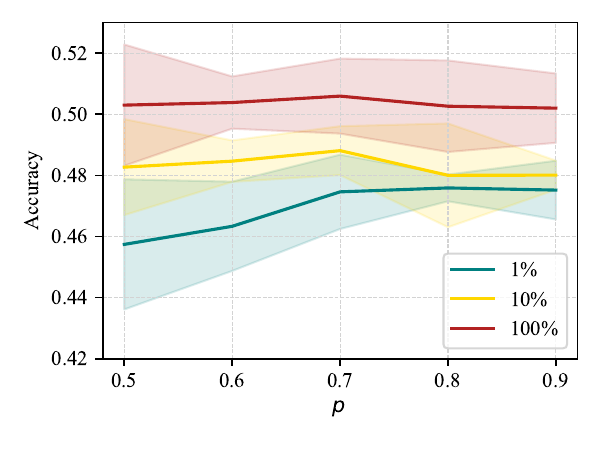}
	}\hspace{-5mm}
	\subfloat[]{
		\includegraphics[width=0.3\textwidth]{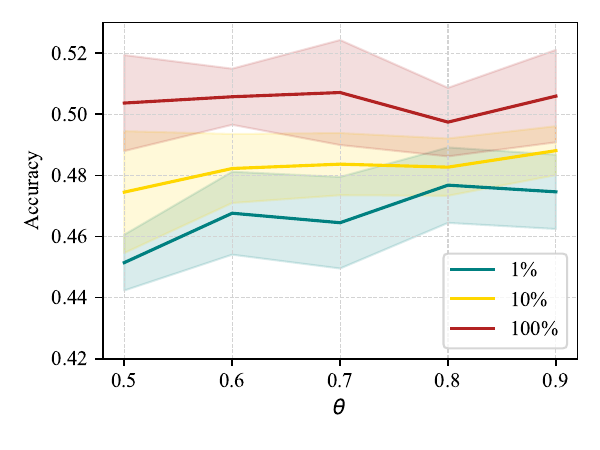}
	}

        \hspace{-5mm}
	\subfloat[]{
		\includegraphics[width=0.3\textwidth]{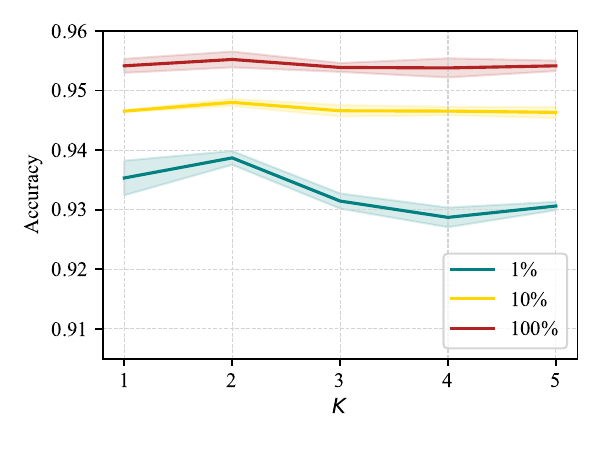}
	}\hspace{-5mm}
	\subfloat[]{
		\includegraphics[width=0.3\textwidth]{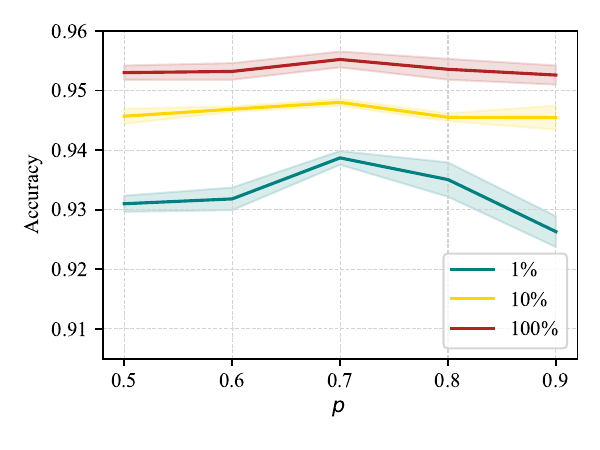}
	}\hspace{-5mm}
	\subfloat[]{
		\includegraphics[width=0.3\textwidth]{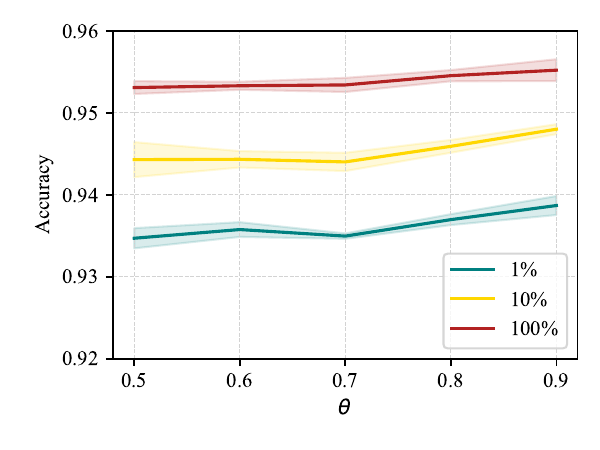}
	}
	\centering
	\caption{Sensitivity analysis results of node classification accuracy for the generation times $K$, unmasking probability $p$, and confidential threshold $\theta$ on the Flickr (a-c) and Reddit (d-f) datasets. Different colors indicate the labeling rates of each dataset, and shading indicates the 95\% confidence interval based on 3 independent runs.\label{sensitivity}}
\end{figure*}

\subsection{Ablation Study}
We further conduct an ablation study to validate the effectiveness of the different components of our model for graph representation learning. The experimental results on the two datasets are presented in Table~\ref{ablation_flickr} and \ref{ablation_reddit}, respectively, where ``w/o feature'' indicates eliminating the feature reconstruction loss, ``w/o mask'' indicates generating pseudo labels with unmasked graphs, ``w/o pseudo'' indicates only using true labels for input and reconstruction, and ``w/o label'' indicates eliminating both true and pseudo label features for input and only using true labels for reconstruction. The results show that the full SLA-VGAE can outperform all variants, demonstrating that the proposed VGAE framework trained by reconstructing the augmented node labels and features is effective in improving the model performance for inductive graph learning. In addition, the superiority of SLA-VGAE over the ``w/o pseudo'' and ``w/o label'' variants becomes larger as the labeling rates of the datasets get smaller, which verifies that the proposed SLAM for label augmentation using self-generated pseudo labels can considerably alleviate the label scarcity problem under weakly supervised learning settings.

\subsection{Sensitivity Analysis}
We also conduct sensitivity analysis on three important hyperparameters related to the proposed SLAM for label augmentation, i.e., the generation times $K$, unmasking probability $p$, and confidential threshold $\theta$. The experimental results are presented in Fig.~\ref{sensitivity}, which demonstrate that the performance of our model for node classification under different labeling rates is relatively robust to all of the three hyperparameters. Specifically, our model performs best when $K=2$. As the generation time becomes larger, the results tend to become stable. Furthermore, the model performance reaches the peak when the unmasking probability $p$ is around $0.7$. A too small value of $p$ will reduce the confidence of the generated pseudo labels, while a too large value will hurt the model generalizability for inferring unseen graph structures. Last, the classification accuracy generally continues increasing as the threshold $\theta$ approaches $1$, verifying the necessity of higher pseudo-label confidence for improving the model performance.

\section{Conclusion}
In this paper, we propose the SLA-VGAE model for semi-supervised graph representation learning. Our model consists of a GCN encoder for node representation learning by performing neighbor aggregation, and a label reconstruction decoder for model training by minimizing the reconstruction loss regularized with a data-agnostic KL divergence. To leverage the label information within the VGAE framework, our proposed model encodes the node labels as one-hot features and then reconstructs the input label features, instead of the adjacency matrix. In addition, to deal with the scarcity of node labels under the semi-supervised learning settings and boost the model generalizability for inductive learning, we propose SLAM to enhance the label information by generating pseudo node labels with the model itself using a node-wise mask approach. Extensive experimental results on benchmark inductive learning graph datasets demonstrate that our proposed SLA-VGAE model achieves competitive results on node classification with significant superiority under the semi-supervised learning setting.

\bibliographystyle{IEEEtran}
\bibliography{ref}

\begin{thebibliography}{10}
\providecommand{\url}[1]{#1}
\csname url@samestyle\endcsname
\providecommand{\newblock}{\relax}
\providecommand{\bibinfo}[2]{#2}
\providecommand{\BIBentrySTDinterwordspacing}{\spaceskip=0pt\relax}
\providecommand{\BIBentryALTinterwordstretchfactor}{4}
\providecommand{\BIBentryALTinterwordspacing}{\spaceskip=\fontdimen2\font plus
\BIBentryALTinterwordstretchfactor\fontdimen3\font minus
  \fontdimen4\font\relax}
\providecommand{\BIBforeignlanguage}[2]{{%
\expandafter\ifx\csname l@#1\endcsname\relax
\typeout{** WARNING: IEEEtran.bst: No hyphenation pattern has been}%
\typeout{** loaded for the language `#1'. Using the pattern for}%
\typeout{** the default language instead.}%
\else
\language=\csname l@#1\endcsname
\fi
#2}}
\providecommand{\BIBdecl}{\relax}
\BIBdecl

\bibitem{hamilton2017inductive}
W.~Hamilton, Z.~Ying, and J.~Leskovec, ``Inductive representation learning on
  large graphs,'' in \emph{Advances in Neural Information Processing Systems},
  vol.~30, 2017.

\bibitem{leskovec2005graphs}
J.~Leskovec, J.~Kleinberg, and C.~Faloutsos, ``Graphs over time:
  {D}ensification laws, shrinking diameters and possible explanations,'' in
  \emph{ACM SIGKDD International Conference on Knowledge Discovery in Data
  Mining}, 2005, pp. 177--187.

\bibitem{paranjape2017motifs}
A.~Paranjape, A.~R. Benson, and J.~Leskovec, ``Motifs in temporal networks,''
  in \emph{ACM International Conference on Web Search and Data Mining}, 2017,
  pp. 601--610.

\bibitem{weber2019anti}
M.~Weber, G.~Domeniconi, J.~Chen, D.~K.~I. Weidele, C.~Bellei, T.~Robinson, and
  C.~Leiserson, ``Anti-money laundering in {B}itcoin: {E}xperimenting with
  graph convolutional networks for financial forensics,'' in \emph{ACM SIGKDD
  International Conference on Knowledge Discovery and Data Mining}, 2019.

\bibitem{subramanian2005gene}
A.~Subramanian, P.~Tamayo, V.~K. Mootha, S.~Mukherjee, B.~L. Ebert, M.~A.
  Gillette, A.~Paulovich, S.~L. Pomeroy, T.~R. Golub, E.~S. Lander
  \emph{et~al.}, ``Gene set enrichment analysis: {A} knowledge-based approach
  for interpreting genome-wide expression profiles,'' \emph{Proceedings of the
  National Academy of Sciences}, vol. 102, no.~43, pp. 15\,545--15\,550, 2005.

\bibitem{rozemberczki2019gemsec}
B.~Rozemberczki, R.~Davies, R.~Sarkar, and C.~Sutton, ``{GEMSEC}: {G}raph
  embedding with self clustering,'' in \emph{IEEE/ACM International Conference
  on Advances in Social Networks Analysis and Mining}, 2019, pp. 65--72.

\bibitem{kipf2017semi}
T.~N. Kipf and M.~Welling, ``Semi-supervised classification with graph
  convolutional networks,'' in \emph{International Conference on Learning
  Representations}, 2017.

\bibitem{velivckovic2018graph}
P.~Veli{\v{c}}kovi{\'c}, G.~Cucurull, A.~Casanova, A.~Romero, P.~Lio, and
  Y.~Bengio, ``Graph attention networks,'' in \emph{International Conference on
  Learning Representations}, 2018.

\bibitem{xu2020inductive}
D.~Xu, C.~Ruan, E.~Korpeoglu, S.~Kumar, and K.~Achan, ``Inductive
  representation learning on temporal graphs,'' \emph{International Conference
  on Learning Representations}, 2020.

\bibitem{zeng2020graphsaint}
H.~Zeng, H.~Zhou, A.~Srivastava, R.~Kannan, and V.~Prasanna, ``Graph{SAINT}:
  {G}raph sampling based inductive learning method,'' in \emph{International
  Conference on Learning Representations}, 2020.

\bibitem{ciano2022inductive}
G.~Ciano, A.~Rossi, M.~Bianchini, and F.~Scarselli, ``On
  inductive--transductive learning with graph neural networks,'' \emph{IEEE
  Transactions on Pattern Analysis and Machine Intelligence}, vol.~44, no.~2,
  pp. 758--769, 2022.

\bibitem{huang2022graph}
Z.~Huang, Y.~Tang, and Y.~Chen, ``A graph neural network-based node
  classification model on class-imbalanced graph data,'' \emph{Knowledge-Based
  Systems}, vol. 244, p. 108538, 2022.

\bibitem{anghinoni2023transgnn}
L.~Anghinoni, Y.-t. Zhu, D.~Ji, and L.~Zhao, ``Trans{GNN}: {A} transductive
  graph neural network with graph dynamic embedding,'' in \emph{International
  Joint Conference on Neural Networks}, 2023, pp. 1--8.

\bibitem{cavallo2023gcnh}
A.~Cavallo, C.~Grohnfeldt, M.~Russo, G.~Lovisotto, and L.~Vassio, ``{GCNH}: {A}
  simple method for representation learning on heterophilous graphs,'' in
  \emph{International Joint Conference on Neural Networks}, 2023, pp. 1--8.

\bibitem{kipf2016variational}
T.~N. Kipf and M.~Welling, ``Variational graph auto-encoders,'' in \emph{NIPS
  Workshop on Bayesian Deep Learning}, 2016.

\bibitem{grover2019graphite}
A.~Grover, A.~Zweig, and S.~Ermon, ``Graphite: {I}terative generative modeling
  of graphs,'' in \emph{International Conference on Machine Learning}, 2019,
  pp. 2434--2444.

\bibitem{mehta2019stochastic}
N.~Mehta, L.~C. Duke, and P.~Rai, ``Stochastic blockmodels meet graph neural
  networks,'' in \emph{International Conference on Machine Learning}, 2019, pp.
  4466--4474.

\bibitem{sarkar2020graph}
A.~Sarkar, N.~Mehta, and P.~Rai, ``Graph representation learning via ladder
  gamma variational autoencoders,'' in \emph{AAAI Conference on Artificial
  Intelligence}, vol.~34, no.~04, 2020, pp. 5604--5611.

\bibitem{tan2022mgae}
Q.~Tan, N.~Liu, X.~Huang, R.~Chen, S.-H. Choi, and X.~Hu, ``{MGAE}: {M}asked
  autoencoders for self-supervised learning on graphs,'' 2022.

\bibitem{hou2022graphmae}
Z.~Hou, X.~Liu, Y.~Cen, Y.~Dong, H.~Yang, C.~Wang, and J.~Tang, ``Graph{MAE}:
  {S}elf-supervised masked graph autoencoders,'' in \emph{ACM SIGKDD Conference
  on Knowledge Discovery and Data Mining}, 2022, pp. 594--604.

\bibitem{guo2022graph}
L.~Guo and Q.~Dai, ``Graph clustering via variational graph embedding,''
  \emph{Pattern Recognition}, vol. 122, p. 108334, 2022.

\bibitem{fan2022heterogeneous}
H.~Fan, F.~Zhang, Y.~Wei, Z.~Li, C.~Zou, Y.~Gao, and Q.~Dai, ``Heterogeneous
  hypergraph variational autoencoder for link prediction,'' \emph{IEEE
  Transactions on Pattern Analysis and Machine Intelligence}, vol.~44, no.~8,
  pp. 4125--4138, 2022.

\bibitem{li2023maskgae}
J.~Li, R.~Wu, W.~Sun, L.~Chen, S.~Tian, L.~Zhu, C.~Meng, Z.~Zheng, and W.~Wang,
  ``What's behind the mask: {U}nderstanding masked graph modeling for graph
  autoencoders,'' in \emph{ACM SIGKDD International Conference on Knowledge
  Discovery and Data Mining}, 2023.

\bibitem{kingma2013auto}
D.~P. Kingma and M.~Welling, ``Auto-encoding variational bayes,'' 2013.

\bibitem{takahashi2019variational}
H.~Takahashi, T.~Iwata, Y.~Yamanaka, M.~Yamada, and S.~Yagi, ``Variational
  autoencoder with implicit optimal priors,'' in \emph{AAAI Conference on
  Artificial Intelligence}, vol.~33, no.~01, 2019, pp. 5066--5073.

\bibitem{velivckovic2018deep}
P.~Veli{\v{c}}kovi{\'c}, W.~Fedus, W.~L. Hamilton, P.~Li{\`o}, Y.~Bengio, and
  R.~D. Hjelm, ``Deep graph infomax,'' in \emph{International Conference on
  Learning Representations}, 2019.

\bibitem{hassani2020contrastive}
K.~Hassani and A.~H. Khasahmadi, ``Contrastive multi-view representation
  learning on graphs,'' in \emph{International Conference on Machine Learning},
  2020, pp. 4116--4126.

\bibitem{zhang2022graph}
W.~Zhang, Z.~Yin, Z.~Sheng, Y.~Li, W.~Ouyang, X.~Li, Y.~Tao, Z.~Yang, and
  B.~Cui, ``Graph attention multi-layer perceptron,'' in \emph{ACM SIGKDD
  Conference on Knowledge Discovery and Data Mining}, 2022, pp. 4560--4570.

\bibitem{sun2021scalable}
C.~Sun, H.~Gu, and J.~Hu, ``Scalable and adaptive graph neural networks with
  self-label-enhanced training,'' 2021.

\bibitem{zhang2021scr}
C.~Zhang, Y.~He, Y.~Cen, Z.~Hou, W.~Feng, Y.~Dong, X.~Cheng, H.~Cai, F.~He, and
  J.~Tang, ``{SCR}: {T}raining graph neural networks with consistency
  regularization,'' 2021.

\bibitem{gorodkin2004comparing}
J.~Gorodkin, ``Comparing two {K}-category assignments by a {K}-category
  correlation coefficient,'' \emph{Computational Biology and Chemistry},
  vol.~28, no. 5-6, pp. 367--374, 2004.

\end{thebibliography}

\end{document}